\documentclass[letterpaper, 10pt, conference]{ieeeconf}      

\IEEEoverridecommandlockouts                              

\usepackage{graphicx}
\usepackage{subfigure}
\usepackage{amsmath}
\usepackage{amsfonts}
\usepackage{bm}
\usepackage{xcolor}
\usepackage{booktabs}
\usepackage{multirow}
\makeatletter
\let\NAT@parse\undefined
\makeatother
\usepackage{makecell}
\usepackage[colorlinks=true, citecolor=black, linkcolor=black]{hyperref}
\usepackage{cite}
\usepackage[nameinlink,noabbrev]{cleveref}

\title{\LARGE \bf
VTAP Gripper: Synergizing Fingertip Sensing and a Visuo-Tactile Active Palm for Dexterous In-Hand Manipulation
}

\author{
Yuhao Zhou$^{1}$, Sheeraz Athar$^{1,\dagger}$, Zhixian Hu$^{1,\dagger}$, Binghao Huang$^{2}$, Yunzhu Li$^{2}$, Juan Wachs$^{1}$, and Yu She$^{1,*}$%
\thanks{$^{*}$Corresponding author. $\dagger$Equal contribution.}%
\thanks{$^{1}$Edwardson School of Industrial Engineering, Purdue University, West Lafayette, IN, USA. {\tt\small \{zhou1437, sathar, hu934, jpwachs, shey\}@purdue.edu}}%
\thanks{$^{2}$Department of Computer Science, Columbia University, New York, NY, USA. {\tt\small \{binghao.huang, yuzhu.li\}@columbia.edu}}%
\thanks{This work was supported in part by the National Science Foundation under Award 2423068 and Award 2520136; and in part by the U.S. Department of Agriculture under Award 2023-67021-39072 and Award 2024-67021-42878.}
}

\crefname{figure}{Fig.}{Figs.}
\crefname{equation}{Eq.}{Eqs.}
\crefname{section}{Sec.}{Secs.}
\crefname{table}{Table}{Tables}

\begin{document}
\maketitle
\thispagestyle{empty}
\pagestyle{empty}

\begin{abstract}
This paper presents a tactile-reactive gripper that integrates a Visuo-Tactile Active Palm (VTAP) and compliant, reconfigurable fingers equipped with tactile array sensors. The design exploits structured finger-palm synergy and multi-modal perception to achieve both robust grasping and fine manipulation. The actuated bi-modal palm seamlessly combines long-range visual localization with contact-rich tactile feedback, substantially extending the system's manipulation capability. To bridge the embodiment gap between human hand motion and the heterogeneous three-finger structure, we further propose a staged, gesture-conditioned retargeting framework for dexterous teleoperation. Extensive experiments validate the system across a range of challenging tasks: reactive grasping of YCB and fragile objects, in-hand syringe reorientation and plunger actuation, singulation of clustered objects down to 3 mm in diameter, and vision-tactile peg-in-hole insertion. Results demonstrate that high manipulation performance can be achieved through coordinated finger-palm interaction and multi-modal sensing, without resorting to high degrees of freedom anthropomorphic designs. The VTAP gripper and its retargeting framework offer a practical reference architecture for dexterous gripper design, manipulation, and contact-rich data collection in support of learning-based approaches. Project webpage: \footnotesize{\texttt{\href{https://yuhochau.github.io/vtap/}{https://yuhochau.github.io/vtap/}}}.
\end{abstract}

\section{Introduction}
The field of robotics is transitioning from structured industrial deployments to consumer-facing services, where robots must operate in dynamic, semi-structured human environments. In industrial settings, manipulation is typically dominated by predefined pick-and-place operations within controlled work cells. In contrast, real-world human environments involve diverse objects, clutter, and varying contact conditions, requiring robots to perform post-grasp actions such as in-hand reorientation and precise alignment. Parallel grippers remain popular because they are mechanically simple and easy to control, and learning-from-demonstration methods such as diffusion policy \cite{chi2025diffusion} further motivate scalable manipulation hardware. However, the limited reconfiguration of parallel grippers restricts dexterous tasks in human-centered environments. Anthropomorphic hands provide higher dexterity but are often bulky, expensive, and difficult to control \cite{wilson2020design}. Multi-finger grippers aim to balance actuation complexity and dexterity, especially when augmented with tactile feedback to expand manipulation capabilities, yet several challenges remain in this area.


\begin{figure}[t]
\vspace{0.2cm}
\centering
\includegraphics[width=0.485\textwidth]{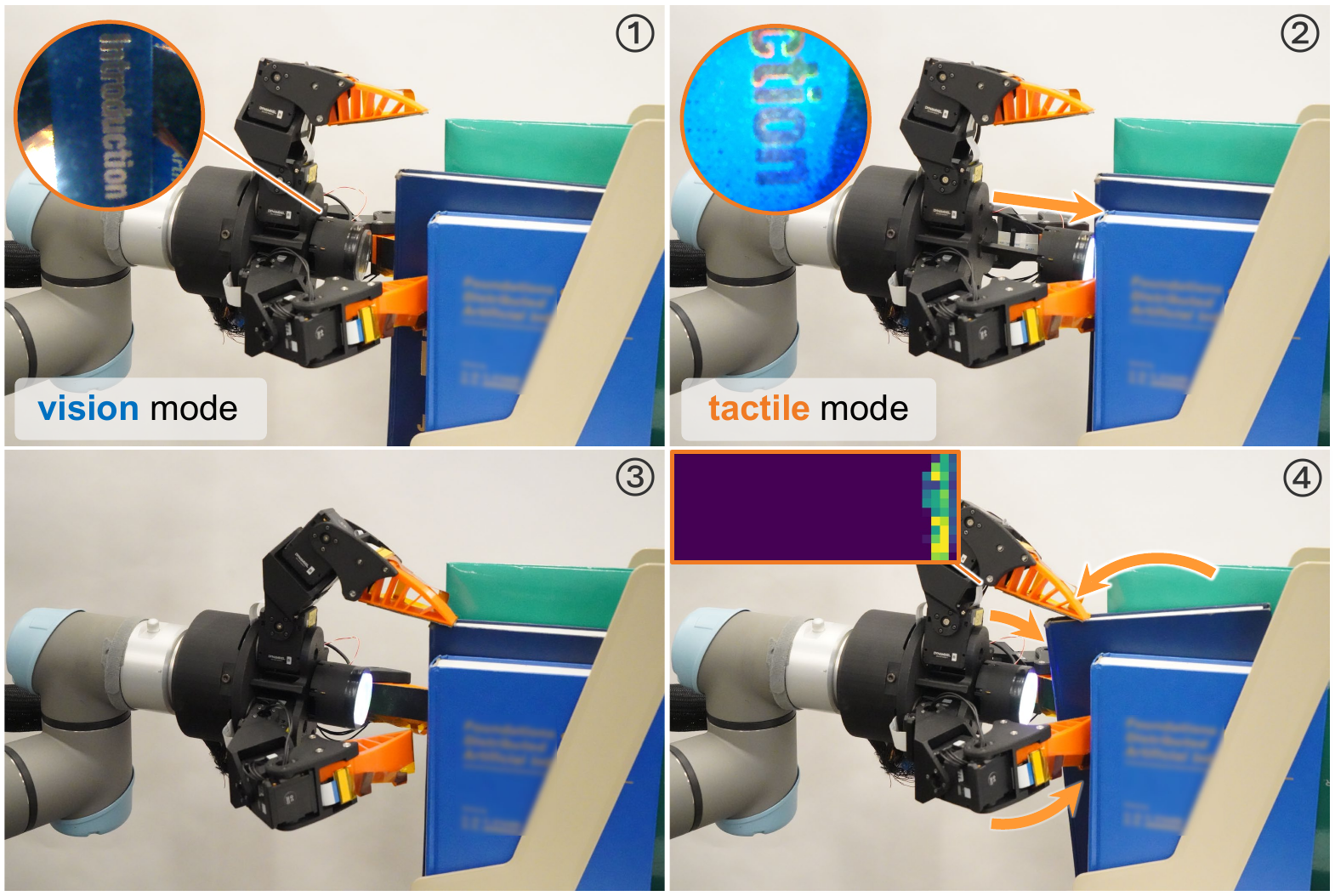}
\caption{Overview of the proposed gripper. The system enables in-hand book insertion into a tightly packed shelf (State~1--2) using the actuated palm with Visuo-Tactile sensing, followed by dexterous in-hand extraction with contact-rich feedback from the fingertip tactile arrays (State~3--4).}
\label{fig:teaser}
\end{figure}

First, in most existing robotic grippers, sensing and actuation are primarily confined to the fingers. Interactions between the object and the palm are often neglected, despite the fact that palm-object contact plays a critical role in many grasping and manipulation tasks \cite{feix2015grasp, bullock2012hand}. Beyond sensing considerations, the mechanical design of non-anthropomorphic multi-finger grippers rarely exploits additional degrees of freedom (DOFs) at the palm. An active palm addresses this gap by enabling controllable object manipulation, particularly in in-hand manipulation scenarios where coordinated contact redistribution across fingers and palm is essential \cite{pozzi2023actuated}.

Second, many existing robotic grippers rely on a single sensing modality, using either vision or tactile sensing. However, a single modality is often insufficient for achieving comprehensive environmental perception in robotic manipulation. Integrating tactile and visual sensing within a unified sensing framework can effectively mitigate occlusion issues while enhancing perceptual robustness. Nevertheless, compactly incorporating multiple sensing modalities, including fingertip tactile sensing, palm tactile sensing, and visual perception, into a single gripper design remains a significant challenge. Addressing this can substantially expand the functional scope and application potential of robotic grippers.

In this work, we present the VTAP Gripper (Fig.~\ref{fig:teaser}), with the following contributions:
\begin{itemize}
\item We propose a tactile-reactive gripper equipped with an actuated bi-modal palm that integrates vision and touch. The design enables structured finger-palm synergy, supporting fine-grained in-hand manipulation while providing both long-range perception and contact-rich feedback.
\item We develop a staged, gesture-conditioned retargeting framework tailored to the proposed gripper. The framework reduces embodiment mismatch, ensures stable and accurate teleoperation, and establishes an efficient data collection approach for downstream learning-based dexterous manipulation policies.
\item We perform experiments on diverse objects and manipulation tasks. The results demonstrate the advantages of integrating an actuated bi-modal tactile palm, showing robust grasping performance and enhanced dexterous in-hand manipulation capability.
\end{itemize}

\section{Related Works}
\subsection{Tactile-Reactive Gripper Design}
Recent progress in tactile sensing has enabled finer-grained perception and substantially expanded the application scope of tactile-reactive grippers. For instance, the integration of piezoresistive sensors into robotic fingers has demonstrated strong adaptability in grasping and manipulating objects \cite{lu2022gtac}. Furthermore, recent studies have incorporated vision-based tactile sensors, particularly the GelSight family \cite{yuan2017gelsight}, into robotic grippers to provide high-resolution tactile feedback for manipulation and closed-loop control \cite{zhang2024gelroller}. Such dense contact information is especially critical for challenging dexterous in-hand manipulation tasks, including small/fragile object manipulation \cite{do2023inter, Hu2025, 11513731}, thin and flexible object handling \cite{11024242}, and deformable linear object manipulation \cite{she2021cable, zhou2025hand}. Although these designs are well-suited for inter-finger manipulation, the interaction between the object and the palm is often overlooked.

To enhance manipulation dexterity, several studies have introduced active palm actuation, demonstrating improved in-hand manipulation through controllable contact redistribution and additional degrees of freedom \cite{teeple2021active, pagoli2021soft, zhou2026tactile}. From a sensing perspective, vision-based tactile palms have been proposed to provide richer contact information \cite{zhang2025soft}. More recently, dual-modality palm designs integrating vision and tactile sensing within a single structure have been explored, for example through mechanical mode-switching mechanisms \cite{11348948} or stereo vision systems \cite{11305117}, enabling both pre-contact perception and post-contact tactile feedback. Despite these advances, coordinated sensing across both the fingers and the palm remains underexplored. In particular, compactly integrating multi-modal palm sensing, coordinated finger sensing, and active palm actuation within a unified gripper architecture remains an open challenge.

\subsection{End-effector Teleoperation for Data Collection}
Learning-from-demonstration methods have recently gained significant attention in robotics due to their ability to directly transfer human manipulation skills to robotic systems. Their appeal lies in enabling efficient skill acquisition without extensive manual controller design. For parallel gripper end-effector control during data collection, prior approaches have utilized interfaces such as 3D SpaceMouse devices \cite{chi2025diffusion}, VR controllers \cite{8461249}, and handheld devices to capture human finger motion. These systems typically rely on additional hardware angle encoders to map human joint states to the robot, as demonstrated in platforms such as UMI \cite{chi2024universal}, Gello \cite{wu2024gello}, and Aloha \cite{fu2024mobile}. In contrast, for anthropomorphic robotic hands with four or five fingers and typically 16 or more DOFs, task-space retargeting is widely adopted \cite{wang2024dexcap}. Human hand keypoints captured via VR devices or data gloves are mapped to robot joint configurations through inverse kinematics. Optimization-based methods are commonly employed to regularize finger positions and orientations, minimizing discrepancies between human and robot hands while preserving overall hand shape \cite{11359455}.

However, retargeting for non-anthropomorphic multi-finger grippers, whose DOFs lie between simple parallel grippers and anthropomorphic hands, is more challenging. Their task-driven and non-structured designs often lack direct kinematic correspondence with the human hand. Nevertheless, these grippers provide a compelling balance between structural simplicity and manipulation capability, making them promising platforms for fine in-hand manipulation. Developing effective retargeting strategies for such systems therefore remains an important and underexplored problem.

\section{Method}
\subsection{Gripper Design}

\begin{figure}[t]
\vspace{0.2cm}
\centering
\includegraphics[width=0.48\textwidth]{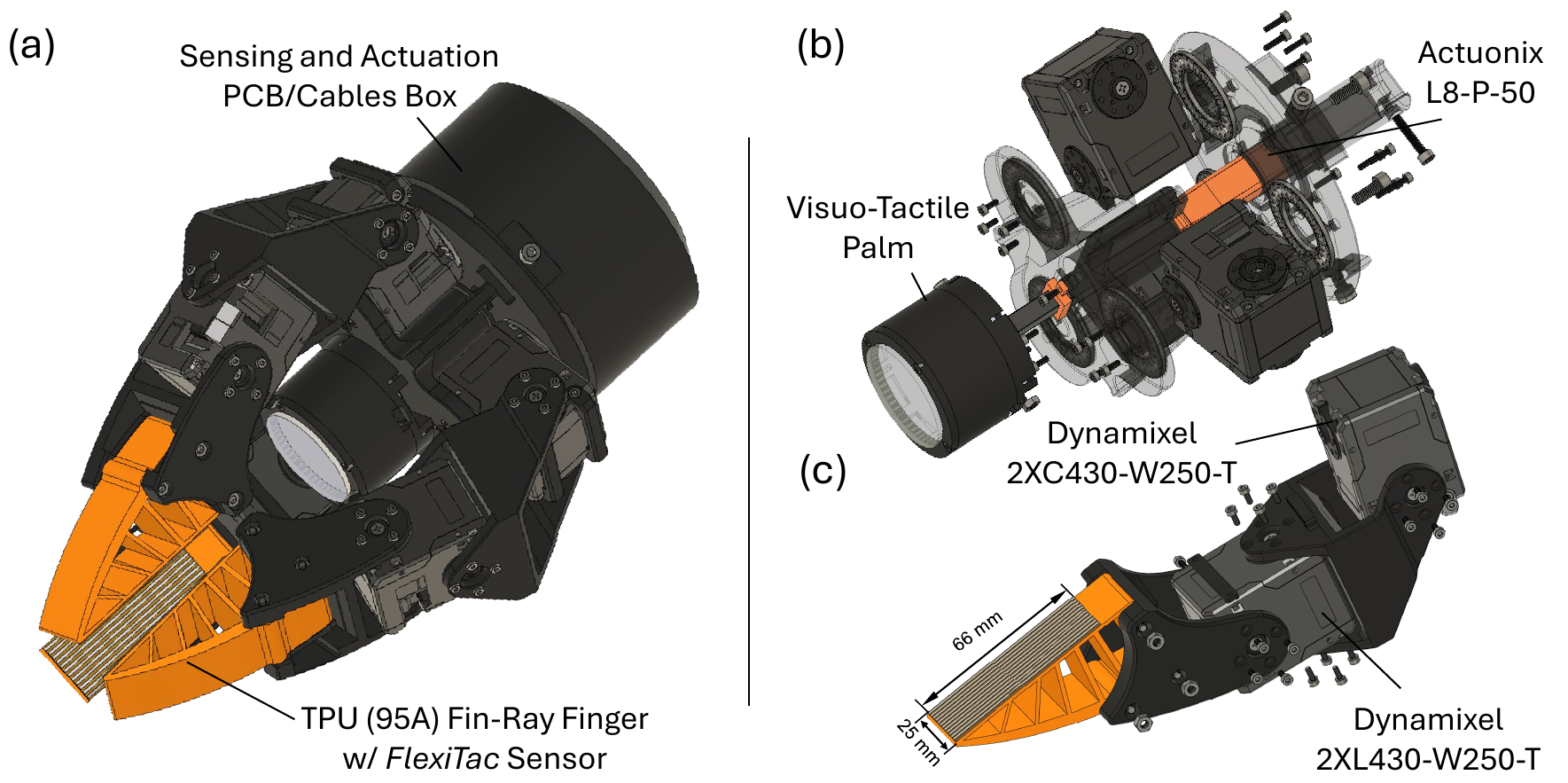}
\caption{(a) Mechanical design and overall structure of the proposed gripper. (b) Active bi-modal palm module. A linear actuator is mounted on the base frame, enabling vertical sliding of the palm. (c) Each finger is actuated by two 2-DOF servo motors. A tactile array sensor is mounted on the fingertip.}
\label{fig:design}
\end{figure}

Fig.~\ref{fig:design}(a) presents an overview of the proposed gripper. The design aims to tightly integrate multi-modal sensing with dexterous actuation, thereby enabling both robust grasping and fine in-hand manipulation. To achieve this objective, an actuated palm module is introduced, as shown in Fig.~\ref{fig:design}(b). A linear actuator (L8-P-50, Actuonix) is mounted within a central slot spanning the upper and lower base frames of the gripper, providing a maximum stroke of $50~\text{mm}$. The actuator output rod is mechanically coupled to the palm, which integrates a vision-tactile bi-modal sensor.

\begin{figure}[t]
\vspace{0.2cm}
\centering
\includegraphics[width=0.49\textwidth]{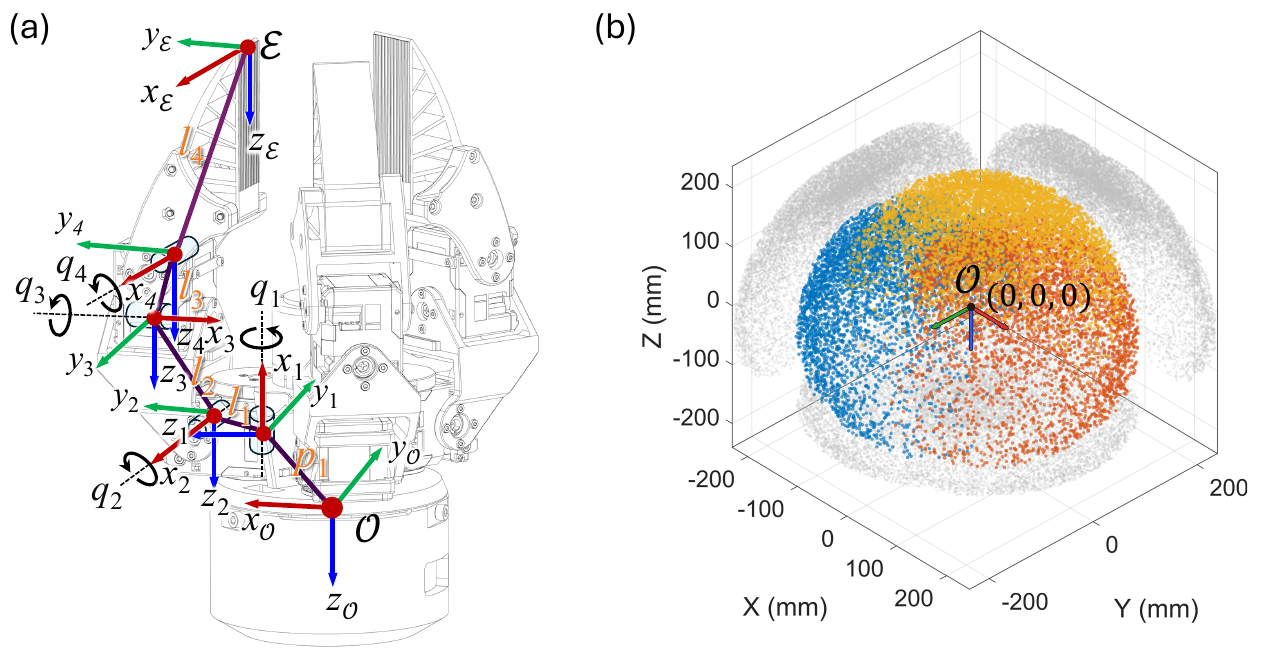}
\caption{(a) Schematic representation of the finger kinematics. (b) Workspace of all three fingertips.}
\label{fig:kinematics}
\end{figure}

As illustrated in Fig.~\ref{fig:design}(c) and with the coordinate definition in Fig.~\ref{fig:kinematics}(a), the gripper consists of three fully actuated and kinematically identical fingers. Each finger is driven by two servo motors, each providing 2 DOFs. A Dynamixel 2XC430-W250-T motor controls radial/ulnar deviation ($q_1$) within $\pm 60^\circ$, and its secondary axis provides metacarpophalangeal (MCP) flexion-extension ($q_2$), corresponding to the proximal joint analogous to the human knuckle, from $-25^\circ$ to $+90^\circ$. A Dynamixel 2XL430-W250-T motor enables abduction/adduction primitives ($q_3$) within $\pm 55^\circ$, while the final actuation axis provides proximal interphalangeal (PIP) flexion-extension ($q_4$), from $-60^\circ$ to $+90^\circ$. The finger is 3D printed in thermoplastic polyurethane (TPU) with a Fin-Ray structure, introducing passive compliance to adapt to objects of varying geometries without explicit contact modeling. Together with an additional 1-DOF active palm, the system provides a total of 13 degrees of freedom.

To formally analyze the manipulation capability of the proposed design, we derive the kinematic model of a single finger, as shown in Fig.~\ref{fig:kinematics}(a). For theoretical analysis, the finger is modeled as a rigid serial chain, and the midpoint of the fingertip is defined as the contact point. Each finger comprises four revolute joints forming a kinematic chain from the base frame $\mathcal{O}$ to the fingertip frame $\mathcal{E}$, with joint configuration defined as $\bm{q} = [q_1, q_2, q_3, q_4]^\top$. The forward kinematics from $\mathcal{O}$ to $\mathcal{E}$ is expressed as:
\begin{equation}
{}^{\mathcal{E}}_{\mathcal{O}}\bm{T}(\bm{q}) =
{}^{1}_{\mathcal{O}}\bm{T}
{}^{2}_{1}\bm{T}
{}^{3}_{2}\bm{T}
{}^{4}_{3}\bm{T}
{}^{\mathcal{E}}_{4}\bm{T}
=
\begin{bmatrix}
\bm{R}(\bm{q}) & \bm{p}(\bm{q}) \\
\bm{0}_{1\times 3} & 1
\end{bmatrix},
\end{equation}
where $\bm{R}(\bm{q}) \in SO(3)$ and $\bm{p}(\bm{q}) \in \mathbb{R}^3$ denote the fingertip orientation and position, respectively. The individual transformation matrices are defined as:
\begin{equation}
\begin{cases}
{}^{1}_{\mathcal{O}}\bm{T} = \bm{D}_{\bm{p}_{1}} \bm{R}_{Z}(\phi) \bm{R}_{Y}\left(\dfrac{\pi}{2}\right) \bm{R}_{X}(-q_1) \\[6pt]
{}^{2}_{1}\bm{T} = \bm{D}_{\bm{l}_{1}} \bm{R}_{Z}\left(-\dfrac{\pi}{2}\right) \bm{R}_{X}\left(\dfrac{\pi}{2} + q_2\right) \\[6pt]
{}^{3}_{2}\bm{T} = \bm{D}_{\bm{l}_{2}} \bm{R}_{Z}\left(-\dfrac{\pi}{2}\right) \bm{R}_{X}(-q_3) \\[6pt]
{}^{4}_{3}\bm{T} = \bm{D}_{\bm{l}_{3}} \bm{R}_{Z}\left(\dfrac{\pi}{2}\right) \bm{R}_{X}(q_4) \\[6pt]
{}^{\mathcal{E}}_{4}\bm{T} = \bm{D}_{\bm{l}_{4}}
\end{cases}
\end{equation}
where $\bm{D}_{\bm{l}_i}$ represents translation along the fixed link vector $\bm{l}_i$, and $\bm{R}_{X}$, $\bm{R}_{Y}$, and $\bm{R}_{Z}$ denote rotations about the corresponding coordinate axes. $\bm{p}_1$ denotes the base offset, and $\phi$ represents the initial azimuthal orientation of each finger's first joint about the base frame. The three fingers are arranged around the palm, each spanning a 120° sector: $[-60^\circ, 60^\circ]$, $[60^\circ, 180^\circ]$, and $[180^\circ, 300^\circ]$. 

Based on the forward kinematics, the reachable workspace of the gripper is estimated via Monte Carlo sampling within the joint limits. Specifically, $5000$ random joint configurations are generated for each joint within its feasible range, and the corresponding fingertip positions are computed and visualized in a 3D coordinate system. The aggregated point cloud approximates the workspace envelope, as shown in Fig.~\ref{fig:kinematics}(b), demonstrating broad and effective spatial coverage around the base frame. For the fingers, the workspace spans approximately $469\,\text{mm}$ along the $X$-axis, $477\,\text{mm}$ along the $Y$-axis, and $311\,\text{mm}$ along the $Z$-axis.



\subsection{Multi-Modal Perception}
\textbf{Fingertip tactile arrays:} to provide rich tactile feedback at the fingertips while maintaining mechanical compliance suitable for Fin-Ray fingers, we integrate thin piezoresistive \emph{FlexiTac} sensors \cite{huang2025vt,huang2026flexitac}. The sensors provide a spatial resolution of $32 \times 12$ taxels over an active sensing area of $66 \times 25\,\text{mm}^2$. Variations in electrical resistance under applied pressure transduce mechanical stimuli into measurable electrical signals, yielding an effective spatial resolution of approximately $2 \times 2\,\mathrm{mm}^2$ per sensing unit.
\begin{figure}[t]
\centering
\includegraphics[width=0.49\textwidth]{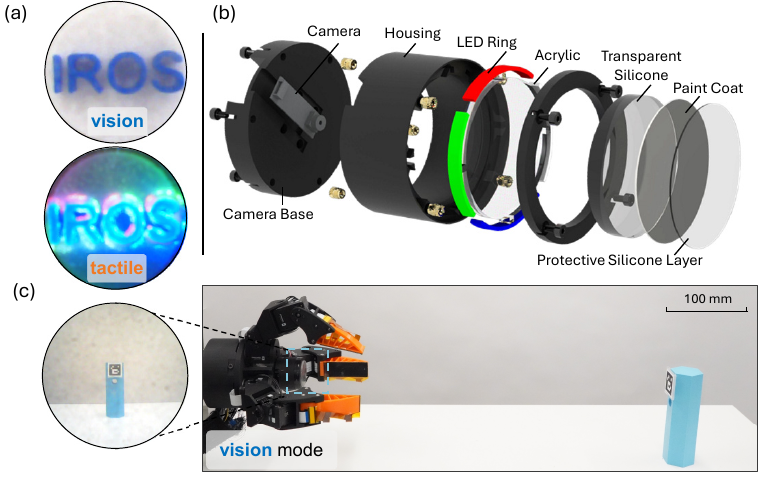}
\caption{(a) Visualization of vision and tactile outputs from the proposed bi-modal palm. (b) Mechanical design of the actuated palm module. (c) Experimental demonstration of long-range visual perception, detecting a $30 \times 30\,\text{mm}$ ArUco marker.}
\label{fig:palm}
\end{figure}

\textbf{Visuo-Tactile bi-modal palm:} in addition to the fingertip sensing, the system incorporates a multi-modal palm perception module. The overall design and representative sensing outputs under the two modalities are illustrated in Fig.~\ref{fig:palm}(a) and (b). To achieve a relatively large depth of field while enabling simultaneous observation of near-field tactile deformations and far-field environmental context, a USB camera sensor (GC0307, GalaxyCore) with a field of view of $50^\circ$ was selected after experimental validation. This dual capability allows the palm to function both as a conventional vision sensor for global scene perception (Fig.~\ref{fig:palm}(c)) and as an optical tactile sensor during contact-rich interactions.

From bottom to top, the camera is enclosed within a $50\,\text{mm}$ diameter cylindrical housing, whose distal end holds a detachable sensing cartridge providing a $40\,\text{mm}$ diameter sensing area. A circumferential ring of multi-color LEDs (red, green, blue, and white) provides controllable internal illumination based on diffuse reflection. The sensing cartridge consists of a $4\,\text{mm}$ thick acrylic substrate supporting a $5\,\text{mm}$ thick silicone elastomer layer. The outer surface of the transparent silicone is coated with mirror-effect paint and covered by a thin protective silicone layer. This reflective coating enables controllable optical transparency: when the LEDs are off, the module remains transparent for external scene imaging; when activated, the illumination is reflected toward the deformable surface, transforming the module into an optical tactile sensor that captures surface deformation. This mechanism enables seamless switching between vision-based perception and tactile sensing without mechanical reconfiguration or stereo camera systems~\cite{11348948,11305117}.

\subsection{Retargeting for Three-Finger Gripper}
\begin{figure}[!t]
\vspace{0.2cm}
\centering
\includegraphics[width=0.48\textwidth]{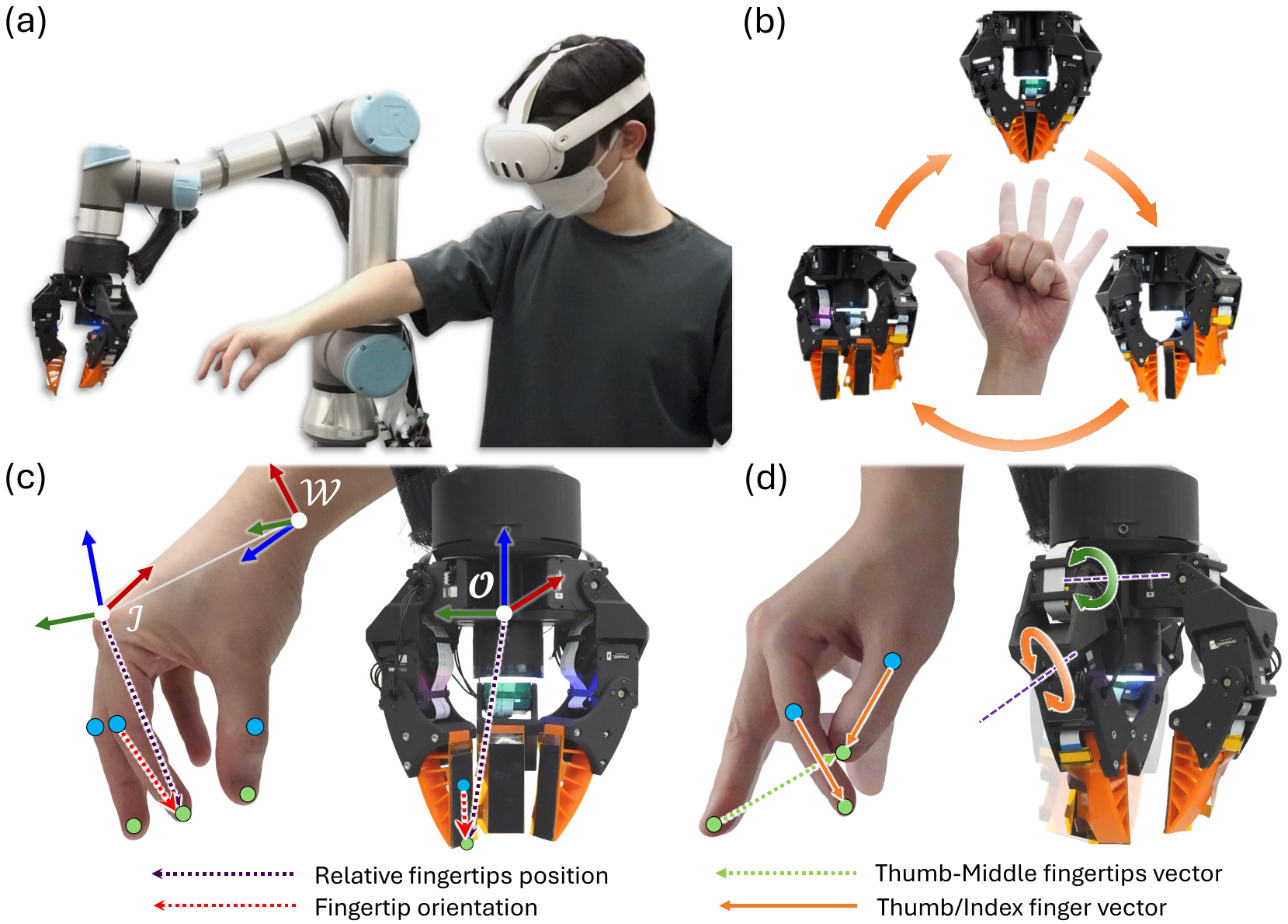}
\caption{Overview of the proposed retargeting framework for teleoperated dexterous manipulation. (a) Experimental setup. (b) Gesture recognition for selecting grasp primitives of the reconfigurable three-finger gripper (cage, power, and pinch). (c) Key optimization objectives for human-to-gripper retargeting. (d) Mapping of thumb-index rolling motions to the gripper’s adduction/abduction DOF for fine in-hand singulation.}
\label{fig:retarget}
\end{figure}

Non-anthropomorphic grippers face a substantial embodiment gap that prevents the direct transfer of human-to-robot retargeting methods. While anthropomorphic hands offer a natural joint-to-joint correspondence, our three-finger gripper exhibits distinct kinematic and geometric challenges. 
First, differences in kinematic structure and joint ordering lead to substantial workspace discrepancies, preventing direct use of anatomy-based optimization objectives; for example, the proposed gripper allows large radial-ulnar deviation beyond human capability. Second, the reference frame definitions differ: the base frame of the three-finger gripper is located at the geometric center of the fingers rather than at a wrist-equivalent location, invalidating wrist-centered vector alignment. To address these issues, we develop a staged control framework for the proposed three-finger gripper, together with task-specific strategies for fine in-hand manipulation, as presented in Fig.~\ref{fig:retarget}.

\textbf{Gesture-conditioned subspace retargeting:} to address structural mismatch, we propose a gesture-conditioned subspace retargeting strategy. Gesture recognition selects grasp priors that constrain radial-ulnar deviation, reducing the configuration space to three representative primitives for three-finger reconfigurable grippers: cage, power, and pinch grasps. Left-hand palm open-close gestures switch between these predefined modes by updating the first joint ($q_1$) of each finger  (Fig.~\ref{fig:retarget}(b)). Furthermore, to preclude kinematic ambiguity arising from the lack of a direct anatomical counterpart, the adduction/abduction DOF is fixed, thereby constraining the subsequent kinematic search space and ensuring physically plausible configurations. To address the base frame challenge, we introduce an intermediate coordinate frame $\mathcal{I}$, defined by applying a constant $SE(3)$ transformation to the VR-captured human wrist coordinate frame $\mathcal{W}$.

After subspace reduction, the remaining MCP and PIP DOFs for each finger are optimized inspired by the established dexterous hand retargeting principles~\cite{11359455}. We use the following primary objectives and present the intuitively crucial factors and definitions in Fig.~\ref{fig:retarget}(c) and (d): 

\subsubsection{Fingertip positions relative to the wrist}
leveraging the intermediate base frame, geometric discrepancies introduced by heterogeneous embodiments are mitigated. For the three principal fingers for controlling the gripper (thumb, index, and middle), we define the fingertip position loss as:
\begin{equation}
\mathcal{L}_{\text{pos}}
=
\sum_{i=1}^{3}
\left\| \mathbf{v}_i^{g} - \mathbf{v}_i^{I} \right\|_2^2,
\end{equation}
where $\mathbf{v}_i^{g}$ denotes the vector from the gripper base frame $\mathcal{O}$ to the $i$-th fingertip, and $\mathbf{v}_i^{I}$ represents the corresponding vector from the intermediate base frame $\mathcal{I}$ to the $i$-th fingertip of the human hand.
\subsubsection{Fingertip orientations}
To enhance orientation observability under VR capture, and given the absence of an equivalent distal interphalangeal (DIP) articulation in the proposed gripper, varying from \cite{11359455, qin2022dexmv}, the human fingertip orientation is defined using the vector from the PIP joint to the fingertip. The corresponding loss is formulated as:
\begin{equation}
\mathcal{L}_{\text{rot}}
=
\sum_{i=1}^{3}
\left\| \mathbf{r}_i^{g} - \mathbf{r}_i^{I} \right\|_2^2,
\end{equation}
where $\mathbf{r}_i^{g}$ represents the vector from the center of the contact plane to the $i$-th robotic fingertip, and $\mathbf{r}_i^{I}$ is the corresponding human PIP-to-fingertip vector. Combining with the smoothness regularization term $\mathcal{L}_{\text{vel}} = \sum_{j=1}^{6} w_j \left\| q_j^{t} - q_j^{t-1} \right\|_2^2$, where $w_j$ represents the joint-specific weight, the overall objective function is formulated as:
\begin{equation}
\mathcal{L}
=
\lambda_1 \mathcal{L}_{\text{pos}}
+
\lambda_2 \mathcal{L}_{\text{rot}}
+
\mathcal{L}_{\text{vel}},
\end{equation}
where $\lambda_1$ and $\lambda_2$ are global scalar weights. To improve numerical stability against tracking noise while preserving sensitivity to small tracking errors, we employ a Huber-type robust penalty function across all internal sub-terms.

Therefore, at each timestep, an optimization problem can be formulated as follows, subject to the joint limits:
\begin{equation}
\mathbf{q}_t^{*}
=
\arg\min
\mathcal{L}(\mathbf{q}_t),
\quad
\mathrm{s.t.}
\quad
\mathbf{q}_{\min}
\le
\mathbf{q}_t
\le
\mathbf{q}_{\max}.
\end{equation} 

\textbf{Singulation retargeting:} we design a retargeting strategy for singulation tasks, which require isolating a single object from multiple clustered items~\cite{zhou2024hand}. Such tasks are representative of dexterous manipulation scenarios and demand precise regulation of fingertip spacing and relative orientation.

As illustrated in Fig.~\ref{fig:retarget}(d), humans accomplish in-hand singulation manipulation through coordinated thumb-index rolling/shearing motions, continuously adjusting both inter-fingertip distance and angular alignment. Inspired by this mechanism, we leverage the reconfigurable pinch mode and the adduction/abduction DOF, mapping only the human thumb and index finger to the two active gripper fingers. Under the gesture-conditioned pinch mode introduced previously, two geometric cues are monitored: the angle between the thumbtip-MCP vector and the index fingertip-PIP vector, and the Euclidean distance between the thumb and index fingertips. When the vectors approach parallel alignment and the fingertip distance falls below a predefined threshold, the controller transitions into singulation mode.

In this mode, a geometric mapping is applied. The human fingertip distance is linearly mapped to equal-magnitude but opposite-direction actuation of the two $q_3$ joints, while the relative angle between the fingertip vectors determines the sign of the motion. Additionally, the flexion of the corresponding MCP joints $q_2$ is regulated through a linear mapping of the thumb-middle distance to enable fine adjustment along the flexion direction, resulting in coordinated control over two active DOFs per finger.

\section{Experiment}
\subsection{Tactile-Reactive Grasping and Manipulation}
We installed the proposed gripper on a UR5e robot arm. To evaluate reactive grasping with fingertip tactile feedback, we adopt a threshold-based stopping criterion using tactile signals for adaptive grasping:
\begin{equation}
Q = \sum_{i=1}^{3} \left[ \alpha \, \Delta C_f + (1-\alpha) \sum_{m,n} S_f(m,n) \right],
\quad
Q > T_{\text{th}},
\end{equation}
where $\Delta C_f$ denotes the change in non-zero contact pixels and $\sum_{m,n} S_f(m,n)$ represents the cumulative signal magnitude of the tactile array taxels on the $i$-th finger. The parameter $\alpha \in [0,1]$ balances contact area and signal intensity. During grasping, each finger continues flexion until $Q$ exceeds an empirically selected termination threshold $T_{\text{th}}$.

\begin{figure}[!t]
\vspace{0.2cm}
\centering
\includegraphics[width=0.43\textwidth]{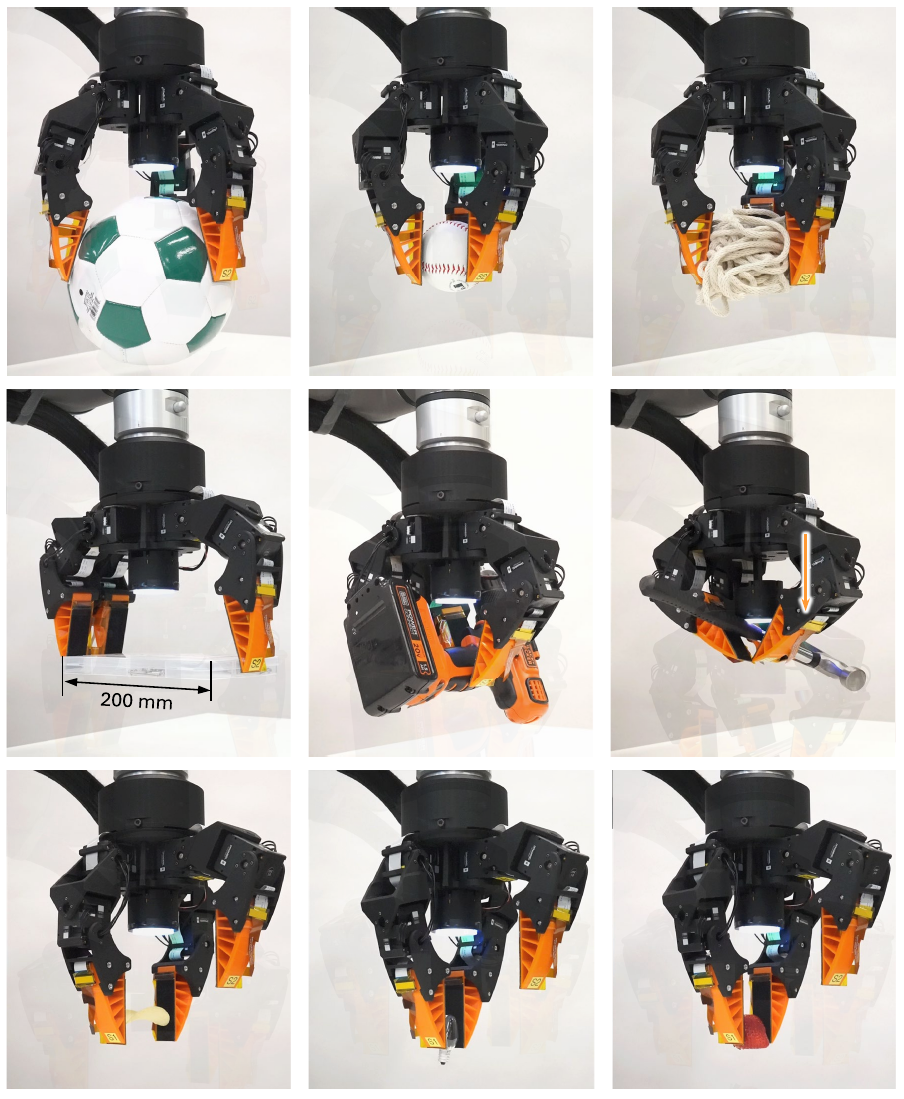}
\caption{Tactile-reactive grasping experiments under three representative configurations: cage grasping of large and articulated objects (top), power grasping of wide and heavy objects (middle), and pinch grasping of fragile objects (bottom).}
\label{fig:grasp}
\end{figure}

\begin{table}[t]
\caption{Grasping Success on Selected Objects Over 10 Trials}
\scriptsize
\centering
\setlength{\tabcolsep}{2.3pt}
\begin{tabular}{c|ccc|ccc|ccc}
\toprule
Objects
& {\scriptsize Soccer} & Baseball & Rope 
& Box & Drill & Hammer 
& \makecell{Potato\\Chip} & \makecell{Light\\Bulb} & Strawberry \\
\midrule
Success & 10/10 & 10/10 & 10/10 & 9/10 & 5/10 & 10/10 & 10/10 & 10/10 & 10/10 \\
\bottomrule
\end{tabular}
\label{tab:manipulation_results}
\end{table}

\begin{figure}[t]
\vspace{0.2cm}
\centering
\includegraphics[width=0.42\textwidth]{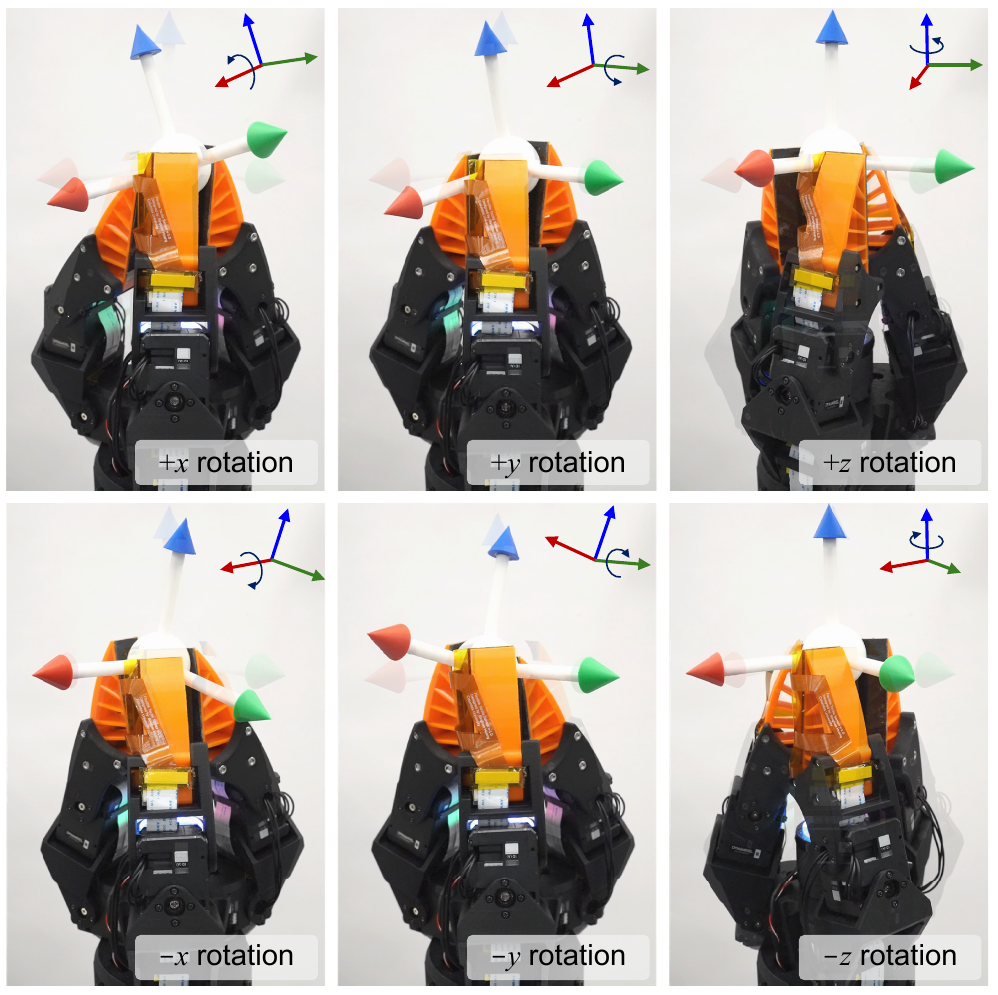}
\caption{In-hand object reorientation experiments.}
\label{fig:rot}
\end{figure}

We evaluate the proposed method on five objects from the YCB object set~\cite{calli2015ycb} together with four additional everyday objects. The objects are grouped into three categories corresponding to the gripper’s representative grasp configurations (cage, power, and pinch, from top to bottom), as illustrated in Fig.~\ref{fig:grasp}. To evaluate grasping performance, we design an experiment that integrates the palm’s visual modality with fingertip tactile sensing: objects are randomly placed within the gripper workspace with arbitrary initial orientations, and the palm camera estimates the object orientation using morphological processing and Canny edge detection, enabling the end-effector pose to be adjusted to an orientation-aligned grasp pose. The tactile-reactive controller is then activated to execute grasping and lifting. A trial is considered successful if the object remains stably grasped throughout the lifting process without dropping.

The results in Table~\ref{tab:manipulation_results} show a high overall grasping success rate of $93.3\%$. The fingertip tactile sensors enabled damage-free grasping of fragile objects, while the power mode and actuated palm supported stable handling of large, wide, and thin cylindrical objects such as the hammer.

Object manipulation experiments were further conducted to evaluate in-hand dexterity. A $40\,\text{mm}$ diameter sphere was rotated about the $x$-, $y$-, and $z$-axes via coordinated finger motions, as shown in Fig.~\ref{fig:rot}. Using predefined orientation trajectories to independently actuate each axis, the object achieved rotations of approximately $\pm 15^\circ$ about the $x$- and $y$-axes and $\pm 20^\circ$ about the $z$-axis. These results demonstrate the gripper’s capability for multi-axis in-hand re-orientation.

\subsection{Syringe Re-orientation and Plunger Actuation}
We validate the effectiveness of the proposed gripper and retargeting framework through a series of in-hand manipulation experiments. At each frame ($\sim25\,\text{Hz}$), hand poses are captured using a Meta Quest 3 VR device, and the corresponding joint configuration $\mathbf{q}_t$ is solved. The resulting commands are temporally upsampled via interpolation to a $100\,\text{Hz}$ control stream. The gripper executes these commands using its built-in current-based PD controller, while the UR5e is controlled via RTDE-based position control.

\begin{figure}[!t]
\vspace{0.2cm}
\centering
\includegraphics[width=0.49\textwidth]{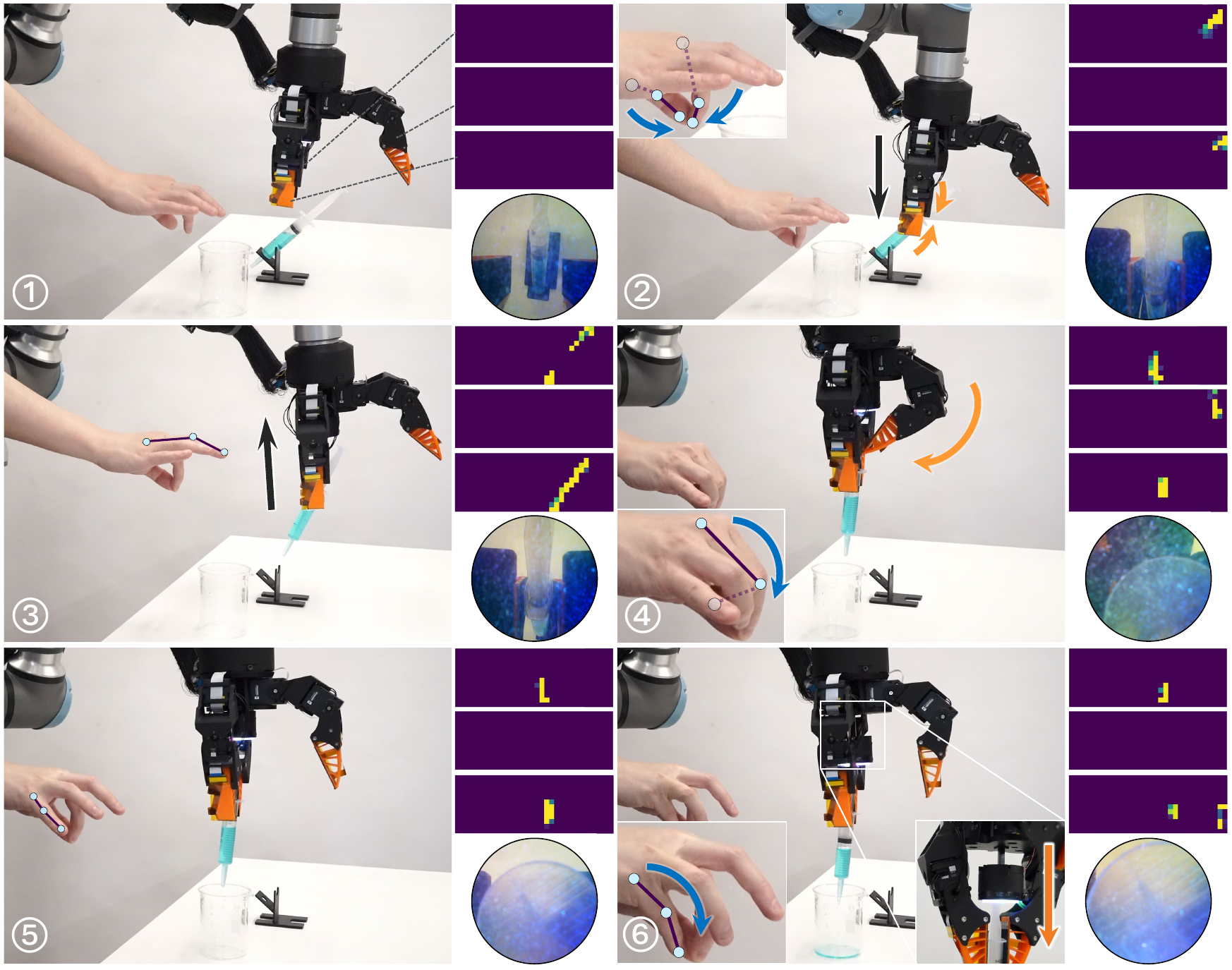}
\caption{In-hand syringe reorientation and plunger actuation via teleoperation using the proposed gripper. The sequence illustrates pinch-based grasping, in-hand reorientation to align the syringe outlet, and coordinated palm actuation for controlled plunger depression.}
\label{fig:syringe}
\end{figure}

The first task evaluating the proposed retargeting strategy is inspired by a laboratory pipetting scenario: extracting a syringe from a stand and depressing the plunger to transfer liquid into a beaker. Experimental snapshots are shown in Fig.~\ref{fig:syringe}. The radial-ulnar DOF is configured for pinch grasping (State~1). When the thumb and index finger close (State~2), the gripper performs a two-finger pinch grasp on the syringe. The tactile arrays capture the contact distribution and in-hand orientation (State~3). Contraction of the middle finger then triggers dexterous in-hand reorientation, aligning the syringe outlet downward (State~4). Finally, contraction of the little finger commands the actuated palm to depress the plunger, completing the liquid transfer (States~5-6).

We use the success rate as the evaluation metric, defined as successful liquid transfer while the syringe remains securely grasped throughout the task. Over 20 consecutive teleoperation trials, the task achieved a success rate of $65\%$ ($13/20$) with an average completion time of $33.4 \pm 5.43\,\text{s}$.
\subsection{In-Hand Singulation Manipulation}
\begin{figure*}[!t]
  \centering
    {\includegraphics[width=1\textwidth]{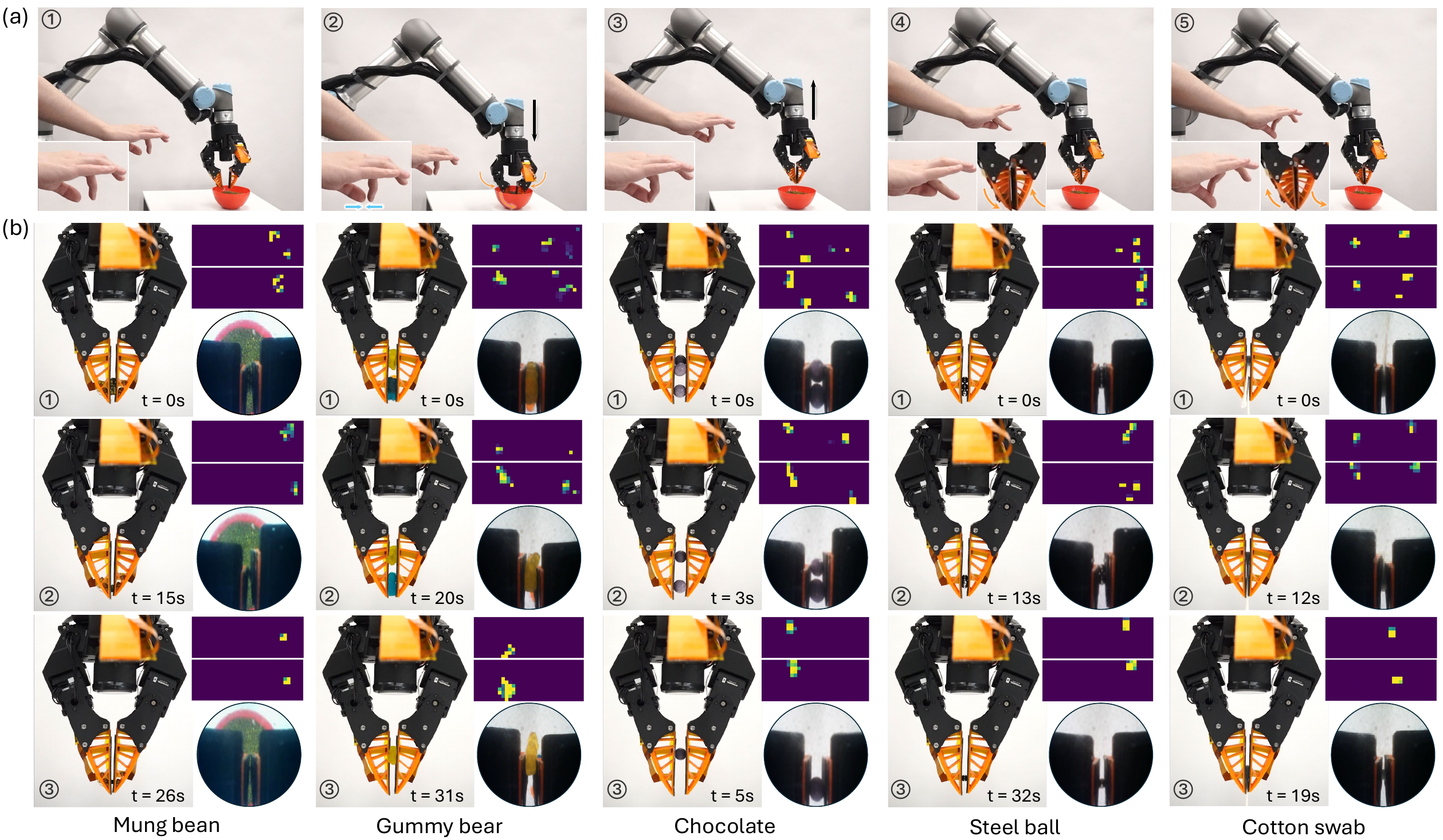}} \\
    \caption{Sequential images of in-hand singulation. (a) Overview of the teleoperated singulation process for grasping granular mung beans from a bowl, where the human thumb-index fingertip vectors are retargeted to control the proposed gripper. (b) Singulation results on five representative objects with varying size, geometry, and stiffness. From the first to the last row: initial multi-object grasp with multiple tactile contact patterns, intermediate object removal during in-hand manipulation, and successful isolation of a single object verified by a distinct tactile contact pattern on each fingertip.}
	\label{fig:singuexp}
\end{figure*}

The in-hand singulation task requires the proposed gripper to grasp a cluster of objects and progressively isolate a single target through fingertip manipulation \cite{zhou2025hand, zhou2024hand}. As shown in Fig.~\ref{fig:singuexp}(a), the gripper first grasps multiple mung beans from a container (States~1--3). During States~4--5, coordinated thumb-index rolling motions are mapped to the gripper’s adduction/abduction DOFs, enabling controlled singulation through in-finger manipulation. To evaluate generality, five target objects with varying size, shape, and stiffness were selected, as illustrated in Fig.~\ref{fig:singuexp}(b): mung beans, soft gummy bear candies, irregular chocolate spheres, $7\,\text{mm}$ diameter steel balls, and cotton swabs.

The first row in Fig.~\ref{fig:singuexp}(b) illustrates the initial grasp, where the tactile arrays clearly indicate multiple patterns. In the second row, partial object removal occurs during manipulation. Finally the last row demonstrates successful singulation, with only a single object remaining, evidenced by a distinct tactile contact pattern on each fingertip. In addition to tactile sensing, the bi-modal palm provides visual feedback during in-hand manipulation, enabling additional perception of the grasp state. From a data collection perspective, this integrated multi-modal observation capability offers strong potential for facilitating learning-based policy training, particularly in confined manipulation spaces where conventional waist-mounted cameras are prone to occlusion~\cite{luu2025manifeel}.

Compared with existing tactile-enabled grippers and their teleoperation setups for similar tasks \cite{10898061}, the proposed framework eliminates the need for teleoperated arm or external motion capture systems, while enabling the manipulation of objects with diameters as small as $3\,\text{mm}$. Furthermore, compared with compact vision-based tactile sensors such as GelSight Mini, the fingertip tactile array offers an approximately $215\%$ larger sensing area, facilitating interaction with a broader range of object geometries.
 
\subsection{Multi-Modal Peg-in-hole}
Finally, we design a fully autonomous peg-in-hole task to demonstrate the functionality of the bi-modal actuated palm in an industrial setting. The objective consists of: (1) localizing a hexagonal prism, (2) precisely detecting a $15\,\text{mm}$ diameter hole on its side surface, and (3) completing insertion onto a $14\,\text{mm}$ peg. A $30 \times 30\,\text{mm}$ ArUco marker is mounted on the top surface of the prism for coarse pose estimation, and the peg pose is assumed known.

\begin{figure}[!t]
\centering
\includegraphics[width=0.45\textwidth]{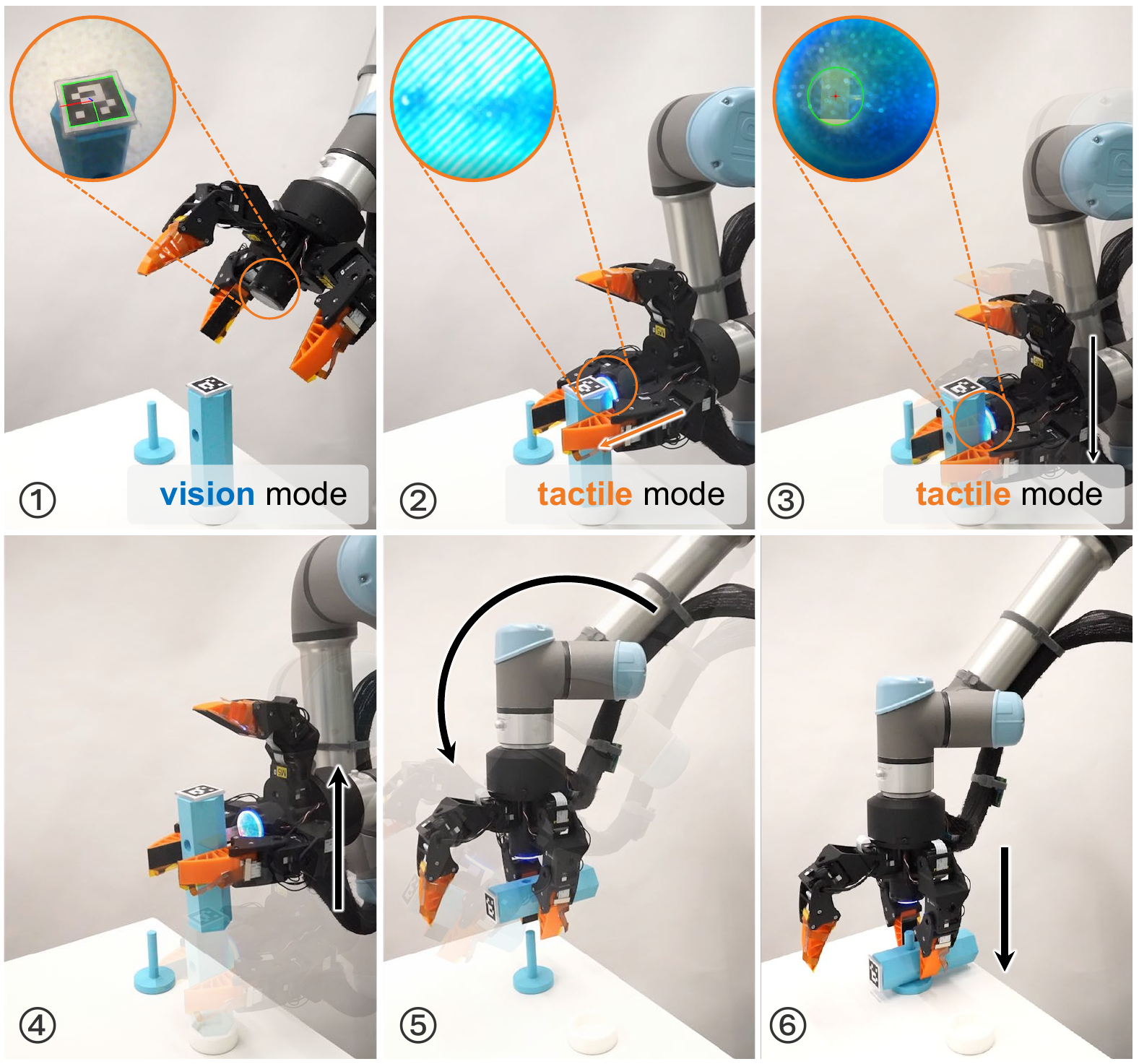}
\caption{Peg-in-hole insertion enabled by the bi-modal actuated palm. Vision mode is used for initial pose estimation, followed by tactile-based hole localization through edge extraction and circle fitting. Finally, the robotic arm adjusts the end-effector pose based on the estimated hole center and executes peg insertion with a $1\,\text{mm}$ tolerance.}
\label{fig:peg}
\vspace{-0.2cm}
\end{figure}

As shown in Fig.~\ref{fig:peg}, during State 1, the palm operates in vision mode to estimate the initial prism pose and guide the grasp. After contact is established, the palm switches to tactile mode by activating internal illumination, and the actuated palm presses against the prism surface to initiate tactile-based hole localization (State 2). The end-effector then moves downward while maintaining contact until the hole center is identified from tactile feedback (State 3). For hole detection, raw tactile images are converted to grayscale and processed using gradient-based edge extraction, followed by contour filtering and algebraic circle fitting. The estimated circle parameters are further refined via outlier rejection to obtain an accurate hole center for in-hand pose correction. Finally, States 4-6 demonstrate successful peg insertion with a tolerance of $1\,\text{mm}$, validating the precision and robustness of the proposed bi-modal palm. We conducted 10 consecutive trials, achieving an overall success rate of $70\%$.

\section{Discussion and Conclusion}
Several issues observed in the experiments merit discussion. In the grasping experiments, the lower success rate for the drill was mainly due to its relatively high weight and asymmetric mass distribution, which occasionally caused slip before the tactile-reactive closure generated sufficient stabilizing contact force. In the syringe task, most failures occurred when the syringe slipped during the retargeted little-finger motion used to depress the plunger. This was mainly caused by VR hand-tracking occlusion: when the headset moved closer to capture the little-finger motion, the thumb or index finger occasionally lost tracking, leading the system to interpret the gesture as finger release and causing the pinch grasp to open unintentionally. This limitation could be mitigated by using data gloves or by leveraging the other hand to control the palm actuation.

In summary, this work introduces the VTAP gripper and its dexterous retargeting framework, demonstrating robust grasping and fine fingertip in-hand manipulation. Experimental results show that high manipulation performance does not necessarily require high-DOF anthropomorphic hands; comparable dexterity can instead be achieved through coordinated finger-palm synergy enabled by the actuated palm. The proposed bi-modal palm provides complementary visual and tactile feedback, benefiting tasks that require multi-modal perception. Future work will investigate integrated finger-palm feedback control to further improve contact-rich manipulation. Furthermore, the retargeting strategy offers a practical reference for controlling non-anthropomorphic three-finger grippers in dexterous manipulation, with potential for scalable data collection.

\section*{Acknowledgment}
The authors would like to thank Zhonghao Mai for helping with the VR hand-keypoint acquisition module.

\bibliographystyle{IEEEtran}
\bibliography{short_bib}

\end{document}